\documentclass[10pt,twocolumn,letterpaper]{article}
\usepackage[utf8]{inputenc}

\usepackage{graphics} 
\usepackage{epsfig} 
\usepackage{mathptmx} 
\usepackage{times} 
\usepackage{amsmath} 
\usepackage{amssymb}  
\usepackage[linesnumbered,ruled,vlined]{algorithm2e}
\usepackage{algpseudocode}
\usepackage{subcaption}
\usepackage{float}
\usepackage{xcolor}

\title{\LARGE \bf Dense Crowd Flow-Informed Path Planning 
}

\author{Emily Pruc, Shlomo Zilberstein, and Joydeep Biswas
}
\date{}

\begin{document}

\maketitle
\thispagestyle{empty}
\pagestyle{empty}

\begin{abstract} 
Both pedestrian and robot comfort are of the highest priority whenever a robot is placed in an environment containing human beings. In the case of pedestrian-unaware mobile robots 
this desire for safety leads to the freezing robot problem
, where a robot confronted with a large dynamic group of obstacles (such as a crowd of pedestrians) would determine all forward navigation unsafe causing the robot to stop in place. In order to navigate in a socially compliant manner while avoiding the freezing robot problem we are interested in understanding the flow of pedestrians in crowded scenarios.
By treating the pedestrians in the crowd as particles moved along by the crowd itself we can model the system as a time dependent flow field. From this flow field we can extract different flow segments that reflect the motion patterns emerging from the crowd. These motion patterns can then be accounted for during
the control and navigation of a mobile robot allowing it to move safely within the flow of the crowd to reach a desired location within or beyond the flow. 

We combine flow-field extraction with a discrete heuristic search to create Flow-Informed path planning (FIPP).
We provide empirical results showing that when compared against a trajectory-rollout local path planner, a robot using FIPP was able not only to reach its goal more quickly but also was shown to be more socially compliant than a robot using traditional techniques both in simulation and on real robots. 

\end{abstract}

\section{INTRODUCTION}
Mobile robots are quickly becoming a routine aspect of daily human life. The increase in public deployment of autonomous mobile robotic systems poses a profound challenge to roboticists: how can we deploy robots in dense human spaces? There are two necessities for this deployment. The robot must be able to plan paths and navigate to goals with optimal efficiency. The second and more challenging necessity is that the robot must maintain pedestrian safety and minimizing disruptions to humans. 
It would seem that this problem simply involves dynamic obstacle avoidance with humans serving as simple obstacles~\cite{art26}. Unfortunately, these solutions often place a cost on collisions that will result in the robot freezing in place for long periods of time rather than risk interacting with a human, this is known as the frozen robot problem~\cite{art18}. Humans are able to avoid such restrictive means of obstacle avoidance through awareness and use of social norms~\cite{art37}. Thus the major challenge facing mobile robot navigation in dense crowds of pedestrians is threefold, path planning 
, obstacle avoidance, while maintaining and utilizing the social rules that inform pedestrian behaviors in crowded scenarios. 

State of the Art solutions to the problem of mobile robot navigation focus on pedestrians as individuals. Due to the chaotic and individualistic behavior of human beings this approach leads to huge variance between scenarios~\cite{art43}. This variance makes it difficult for social navigation models to be generalizable as all solutions must either be tuned for or trained for specific scenarios~\cite{art28,art47}. This has created a tendency towards bespoke solutions, where a single environment is selected and optimized for such as offices or airports~\cite{art44,art45}. In order to achieve socially-acceptable and efficient performance of mobile robot in dense crowds across all scenarios we propose a method of treating dense crowds of pedestrians as particles suspended in a fluid. This will allow us to extract the overall flow of a crowd and apply the same rules that govern the motion of the crowd to the motion of our robot. Thus, producing compliant robot motion within and around the crowd. 

The major contributions of this work are i) the extraction of crowd flow from 3D LiDAR data by treating pedestrians as particles suspended in a fluid and applying fluid dynamic techniques to ascertain the forces exerted on the pedestrians by the crowd's flow, ii) the use of this extracted flow in the mobile robot path planning algorithm and iii) demonstrating how Flow-Informed path-planning results in more socially compliant robot navigation in dense crowds of pedestrians than do other social navigation methods.

\section{RELATED WORK}

Our work in flow-informed path planning builds on work done in two disparate areas, modeling pedestrian dynamics as fluid flows and mobile robot social navigation. 

\subsection{Modeling Crowds of Pedestrians as Fluid Flows}
The fact that a sufficiently dense crowd of pedestrians moves with many attributes of a fluid has been well-studied since the 1970s, when Henderson began studying crowds as fluids where the motion of the flow was studied using discrete simulations of individual pedestrians\cite{art21}. It was then acknowledged by Hughes that while large crowds do have the attributes of a fluid it is that of a rational thinking fluid modeled by a set of social rules. Thus, the traditional equations of fluid mechanics, such as the Navier-Stokes equation, were not able to adequately model the flow of a human crowd~\cite{art20}. New equations based on the concepts of fluid-mechanics were derived in order to represent the non-linear self-driven continuum that is a large crowd of pedestrians.  

One such emerging equation was adapted from Langevin's equation for approximating the motion of Brownian particles using the fluctuating forces acting on said particles~\cite{art22}. The Langevin equation served as a good starting point to calculate the forces inherent on a pedestrian in a crowd. All that was necessary to complete a Newtonian model of the motion of pedestrians within a crowd flow was the introduction of a driving force that represented the pedestrian's innate desire to reach their goal~\cite{art10}. 

\subsection{Social Navigation}
Seamless and efficient navigation through an environment is consistently a major focus for any mobile robot. This becomes even more true when the robot is introduced into a human environment. Now it is essential for the robot to not only perform seamlessly and efficiently but it must also prioritize pedestrian safety and comfort. Due to the importance and complexity of this task, social navigation is a well studied field. Mobile robot navigation in dense crowds can be broken down into two different approaches, model-based solutions and learning-based solutions.

Model-based solutions explicitly map social interactions to specific states in an attempt to extend multi-agent collision avoidance so that they incorporate social mores~\cite{art23}. Many of these model-based solutions involve repeatedly applying a series of equations to each individual pedestrian in order to predict their motion, because of this model-based solutions cannot easily be applied to large groups and are not generalizable 
to different scenarios. 

The most prevalent and well known model-based solution in the realm of social navigation is that of the Social Force model detailed by Helbing and Moln\'ar~\cite{art23}. The main contribution of the Social Force model is that human trajectories can be explained in terms of social forces. Each of these social forces corresponds to a social rule inherently obeyed by human beings. The first rule is that a pedestrian wants to reach a given destination as comfortably as possible, this results in the goal force. Second, the motion of the pedestrian is influenced by other pedestrians,and the environment(i.e obstacles). Together these components make up the social forces acting on a pedestrian. The social force model has been used in conjunction with path-planning techniques and multi-object optimization techniques to navigate in dense crowds~\cite{art24},~\cite{art25}. 

Human-robot proxemics has also served as an important model for mobile robot social navigation~\cite{art39}. By designing robot proxemic behaviors, physical and psychological distance from others, after human social norms by exploiting social cues like eye gaze robots can ensure navigation in a socially compliant manner~\cite{art40}. 

One drawback of the model-based solutions is that  each pedestrian must be considered and accounted for individually. This makes the model-based methods  cumbersome for large numbers of pedestrians and such a calculation becomes intractable.

The second method of solving robot navigation in crowds of pedestrians is learning-based methods, in which deep learning techniques are used to either predict pedestrian motion or to learn a method of navigation by observing human demonstrations~\cite{art32,art34}. Like the model-based solutions, learning-based solutions consider pedestrians as individuals making these methods impractical for large crowds. Also like model-based solutions the realities of training deep-learning models means most deep-learning solutions are not generalizable. 

Three popular approaches to learning-based solutions for navigation in crowds are deep reinforcement learning, inverse reinforcement learning and supervised-learning. 

Deep reinforcement learning has the advantage of being able to develop a time efficient navigation policy that is also able to employ common crowd social norms. Deep reinforcement learning has been employed to assume the behavior of decision-making agents (pedestrians) and allow for the observations of an arbitrary number of agents, rather than imposing a maximum number of observable agents~\cite{art30}.  In 2018 Chen et.al. presented socially aware collision avoidance with deep reinforcement learning (SA-CADRL) that is able to mimic human passing behaviors and match the speed of humans moving through an environment 
~\cite{art28}. Using reinforcement learning means that a human expert has to design a bespoke cost-function, in an effort to alleviate this necessity inverse reinforcement learning has been employed to generate these cost functions.

Perez-Higueras and colleagues utilized IRL to learn navigation behaviors from human demonstrations and then applied the learned cost function to Optimal Rapidly-exploring Random Trees (RRT*). This allowed for socially compliant robot path planning in human environments~\cite{art32}. A three module IRL based solution was developed by Kim and Pineau; the first step was feature extraction, then IRL was used to learn a set of demonstrated trajectories, and finally a path planning module composed of a global planner, a local planner, and an obstacle avoidance system served as a method of social navigation~\cite{art31}. Although these solutions result in socially acceptable path planning without the need for expert cost-function determination, these methods are not generalizable as the behaviors learned from demonstration are highly domain dependent. 

Supervised-learning is another popular method of ensuring safe and compliant robot navigation. Pedestrian surveillance has been used to allow robots to mimic human motion and predict human velocity in an environment. This allows the robot to consistently reach a goal safely and with social compliance~\cite{art34}. Convolutional neural networks have also been employed to predict a cost map that allows the robot to act similarly to the behavior learned from pedestrian demonstrations~\cite{art27}. Graph convolutional neural networks have been used in conjunction with an attention network that accurately predicts human attention in crowded scenes and then learns a method for crowd navigation~\cite{art29}. Although both of these supervised-learning methods allow for some extensibility in the size of the crowd navigated, they are still constrained by their training data. 

In order to address the generalizability limitation of both model-based and learning-based solutions we propose to use fluid mechanics, which can be applied to any crowd size or shape, to extract the flow of pedestrians and then use said flow to plan a safe and socially compliant path for the robot to reach it goal state.

\section{FLOW-INFORMED PATH PLANNING}

In order to allow for socially-acceptable and efficient mobile robot navigation in dense crowds of pedestrians we propose a Flow-Informed path planning(FIPP) solution that uses the inherent motion of a group of pedestrians to reach a goal location.  

Given observations of human positions and velocities at each time step,  \(H_t = \{h_i\}\), the current location of the robot (\(r\)), a navigation goal (\(goal\)), and an objective cost function. The cost function seeks to minimize the time it takes for the robot to reach the goal while maximizing the socially compliant behavior of the robot (i.e minimize the robots interference on human behavior).  

\begin{equation}\label{OCF}C_{\phi}= \min(C_T + C_F)\end{equation}

Where \(C_{\phi}\) is the total cost of a path \(\phi\) to the goal, \(C_T\) is the total cost of the path in terms of efficiency to goal and \(C_F\) is the social compliance cost. Given these inputs FIPP is able to find a socially-compliant and efficient solution to a given goal from the robot's current position. 

Using the observations of human activity a flow field \(F_{total}\) is extracted for the area currently or previously visible to the robot. This flow-field is then used with objective cost function to find the shorted path to the goal. The Cost \(c_{\phi_i}\) of each action is equal to the sum of distance from \(goal\) of the new state \(c_g\), the distance from \(r\) of the new state \(c_r\), and the cost of moving with respect to the flow field \(c_f\). \[c_{\phi_i}= c_g + c_r + c_f\] The minimum cost path is then extracted using equation \ref{OCF}. 


Before the robot can plan with respect to a flow-field it first must extract information on the flow-field from the movement of the crowd. To do this we apply the Active-Langevin Equation~\cite{art48} to determine the force on a particle caused by the inherent crowd flow. 

\subsection{The Active-Langevin Equation} 
 
 The Active-Langevin equation represents this total force term as a composition of three distinct forces, a frictional force (\(F_\text{friction}\)), a driving force (\(F_\text{driving}\)), and any random force or noise affecting the pedestrian (\(F_\text{random}\)),
 
  \begin{equation}F_\text{total}=F_\text{friction}+F_\text{driving}+F_\text{random} \end{equation}
  
as opposed to Brownian particles which experience both random forces and disturbances with regularity. We assume that human beings do not undergo such variability in influence and thus \(F_\text{random} = 0\). This reduces our total force equation to the sum of the friction and driving forces. 

\subsubsection*{The Friction Force}
In the original definition of the Active-Langevin equation the friction force represents the interaction of a given particle with the particles surrounding it. This frictional force still applies to pedestrians moving in dense crowds, i.e a pedestrian will be slowed down by large numbers of people moving in the opposite direction around it. Thus the frictional force can be represented as 

\begin{equation}F_\text{friction} = -\mu_i v_i(t)\end{equation}

Where \(\mu_i\) is the coefficient of friction of the crowd at each pedestrian's location. This coefficient of friction can be thought of as the viscosity of the pedestrian and its neighbors and thus can be calculated using the relationship between a pedestrian and its neighbors~\cite{art10}. 

\begin{equation}\mu_i = 1 - \frac{1}{j_n\text{max}(|x_i-x_{j_k}:k=1...n|)}\sum^{j_n}_{j=1}|x_i-x_j|\end{equation}

In the calculation above \(j_n\) represents the total number of neighbors surrounding a pedestrian, \(x_i\) represents the position of the ith pedestrian while \(x_j\) represents the position of the jth neighbor of the ith pedestrian. In turn, \(|x_i-x_j|\) represents the distance between two pedestrians.  

\subsubsection*{The Driving Force}
The driving force can be thought of as the sum of the self-propelling force of the pedestrian and the influence force between a pedestrian and their neighbors.  
The driving force of a pedestrian is equal to its velocity times its self-driving coefficient.

\begin{equation}F_\text{driving}=F_\text{influence}+F_\text{self-propelling}\end{equation}

The self-propelling coefficient represents the activity of a pedestrian trying to reach its goal in the easiest and most efficient way possible. The self-propelling coefficient was empirically calculated as 0.5 by Behera et al\cite{art10}. This makes the self-propelling force, 

\begin{equation}F_\text{self-propelling} = \xi v_i(t)\end{equation}

where \(v_i\) is the speed of the current pedestrian. The influence potential is the average interaction of the ith pedestrian with its neighboring particles and can be calculated using an interaction coefficient (\(\alpha\)), the average pedestrian velocity(\(v_{avg}\)) and the relative velocity (\(v_{rel}\))between a pedestrian and its neighbors. 

\begin{equation}F_\text{influence} = \alpha(v_i(t)-v_{rel}(t))\end{equation}

\begin{equation}v_{rel}(t) = \sum_{j=k}^{j_n}v_k(t)\end{equation}

Where \(v_k\) is dependent upon a radius at which a pedestrian will influence another member of the crowd. This radius of influence \(h\) can be adjusted to reflect different crowd densities and behaviors, given the distance \(d_{ik}\) from the ith pedestrian to the kth pedestrian, 

\[ \begin{cases} 
      d_{ik} <= h & v_k = \text{kth pedestrians velocity}\\
      else & v_k = 0
   \end{cases}
\]

The average velocity is calculated with respect to all pedestrians observed at a single time step, 
\begin{equation}v_{avg}(t) = \frac{1}{j_n}\sum_{j=1}^{j_n}v_j(t)\end{equation}

Thus \(\alpha\) can be calculated as, 

\begin{equation}\alpha = \frac{v_{rel}(t)}{v_{avg}(t)}\end{equation}

Finally,  the driving force can be written as, 
\begin{equation}F_\text{driving}=\alpha(v_i(t)-v_{rel}(t))+ \xi v_i(t)\end{equation}

The Active-Langevin equation representing the motion of pedestrians in the crowd 
can be represented as 

\begin{equation}F_\text{total}=F_\text{friction}+F_\text{driving}\end{equation} 
\begin{equation}F_\text{total}= -\mu v_i(t)+\alpha(v_i(t)-v_{rel}(t))+ \xi v_i(t)\end{equation} 

 \begin{figure*}[h!] 
 	\centering
 \begin{subfigure}[t]{1.5in}
		\centering
		\includegraphics[width=0.9\linewidth]{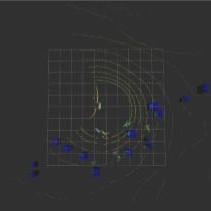}
		\subcaption{}
		\end{subfigure}
\begin{subfigure}[t]{1.5in}
		\centering
		\includegraphics[width=1\linewidth]{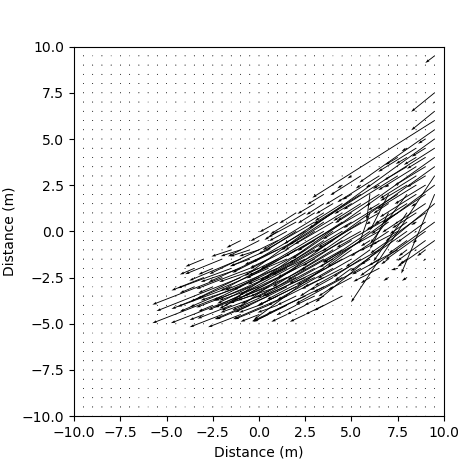}
		\subcaption{}
		\end{subfigure}
\begin{subfigure}[t]{1.5in}
		\centering
		\includegraphics[width=1\linewidth]{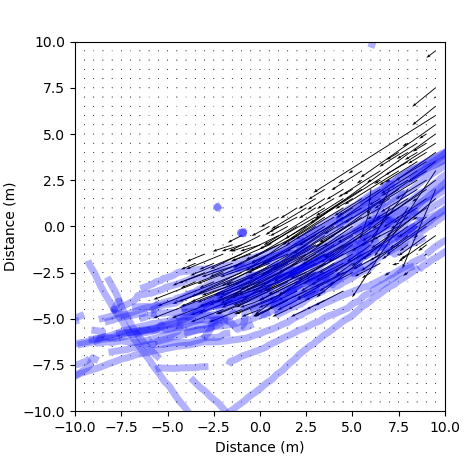}
		\subcaption{}
		\end{subfigure}
\begin{subfigure}[t]{1.5in}
		\centering
		\includegraphics[width=1\linewidth]{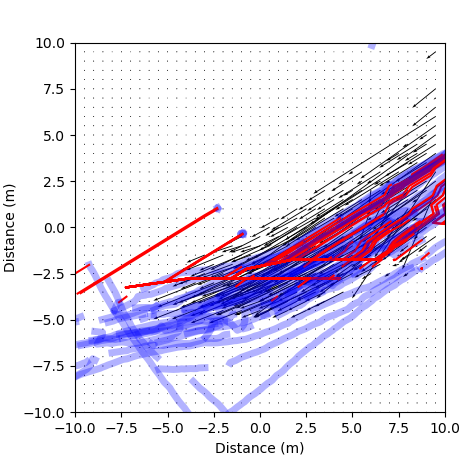}
		\subcaption{}
		\end{subfigure}

\caption{Flow-field verification: a) the tour group scenario, b)the extracted flow field c) blue true pedestrian paths layered on the flow field, d) red flow predicted pedestrian paths. }
\label{fig:fver}
\end{figure*}

\subsection{Flow Extraction} 

The first step in our FIPP system is that of Flow extraction. This is done by imposing a 2D mesh grid on the space visible to the robot's LiDAR. Then as pedestrians enter or move around the space visible to the robot their locations, directions, and velocities are identified using hdl\_people\_tracking, a ROS package for real-time people tracking using 3D LiDAR ~\cite{art16}. The centroid of each pedestrian is then mapped to their closest location on the mesh grid. Once all of the pedestrians in a given time step have been mapped to their locations on the mesh grid, the Active-Langevin equation is applied to every location on the mesh-grid, with one small change. For the ease of calculation we determine a radius of potential influence \(h\), this represents a distance up to which neighbors can exert influence on the a given location. When calculating the relative velocity, necessary for the calculation of the influence force, the velocity of any neighboring location that is further away from the current position than \(h\) is ignored. By computing the Active-Langevin equation at each mesh-grid location we can predict the direction and magnitude of the flow in locations where the pedestrians have either not yet reached or will influence a later pedestrian to visit, we refer to this mesh grid as a flow field.

\subsection{Flow-Informed LAO*}

All though any discrete heuristic search can be used in conjunction with the flow extraction to satisfy the steps of FIPP, we chose to use LAO*~\cite{art41} due to its speed and ability to find an optimal path without needing to evaluate all possible states and actions. In order for LAO* to find an optimal path, not only based on efficiency to goal but also social acceptability based on the flow-field, the heuristic was adjusted. The cost \(c_a\) of each action is equal to the sum of distance from \(goal(x,y)\) of the new state \(c_g\), the distance from \(r(x,y)\) of the new state \(c_r\), and the cost of moving with respect to the flow field \(c_f\). 
\begin{equation}c_a = c_g + c_r + c_f\end{equation}

The flow cost \(c_f\) is determined by the action with respect to the flow-field. When the action results in the robot moving in the same direction as the flow-field the cost will be low and when the robot moves in the opposite direction from the flow-field the cost will be high.

\section{EXPERIMENTAL RESULTS}

We present results from experiments in crowded social scenarios that show i) a verification of the flow-field extraction by its ability  to correctly predict the motion of pedestrians moving through the flow, ii) FIPP results in more efficient navigation in and around crowds than does traditional path planning, iii) flow informed path planning results in more socially-compliant navigation  than does traditional path planning, and iv) the use of FIPP for efficient and compliant navigation in real-world scenarios with real robots.

\begin{figure*}[h!]
\centering
\begin{subfigure}{1.5in}
    \centering
    \includegraphics[width=.95\textwidth]{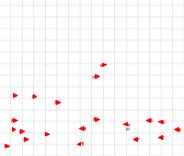}
    \subcaption{Chaotic}
\end{subfigure}%
\begin{subfigure}{1.5in}
    \centering
    \includegraphics[width=.95\textwidth]{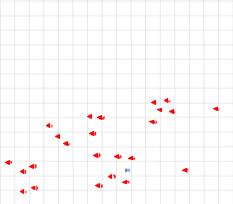}
    \subcaption{Single Flow}
\end{subfigure}
\begin{subfigure}{1.5in}
    \centering
    \includegraphics[width=.95\textwidth]{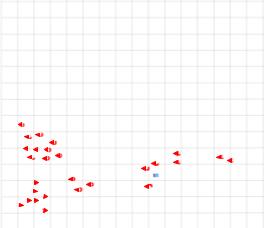}
    \subcaption{Double Flow}
\end{subfigure}%
\begin{subfigure}{1.5in}
    \centering
    \includegraphics[width=.95\textwidth]{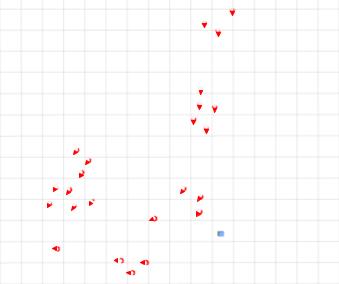}
    \subcaption{Intersection}
\end{subfigure}

\caption[short]{The different simulator test scenarios. Humans and their directions are represented by red triangles, the robot is a blue square}
\label{fig:scenes}
\end{figure*} 

\subsection{Flow-field Extraction Verification} 

To evaluate the validity of our flow-field extraction we observed and extracted flow-fields from a real-world scenario on the University of Massachusetts at Amherst campus, one of a tour group. After extracting the flow field the paths of each pedestrian were recorded based on their locations as identified by the pedestrian tracker. Given the starting point of each pedestrian their theoretical trajectory were calculated by applying the magnitude of direction of the flow at the pedestrians locations to them, thus moving them through the field. This predicted trajectory was compared to the real trajectory of the pedestrian and the average distance between the two trajectories was calculated for each scene. The average distance in trajectory for both scenarios was less than 20cm. Since the flow-field was able to predict the motion of pedestrians moving through it with high accuracy, it is clear that the extracted flow-field is a valid representation of the forces inherent on the pedestrians that make up the crowd (figure \ref{fig:fver}).
\begin{figure*}[h!]
\centering
\begin{subfigure}[t]{2.75in}
		\centering
		\includegraphics[width=1\linewidth]{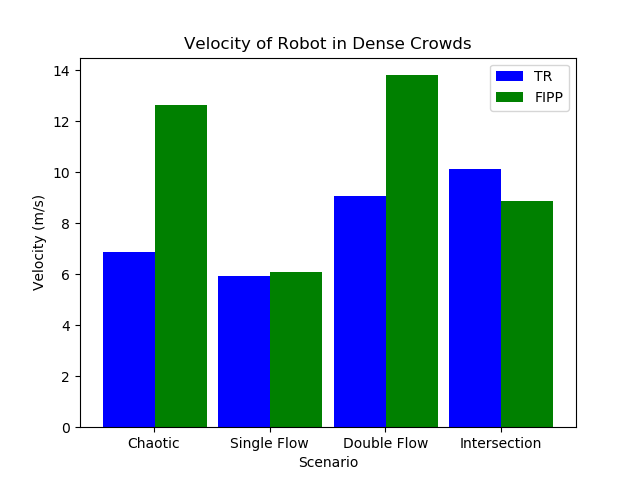}
		\subcaption{}
		\end{subfigure}
\begin{subfigure}[t]{2.75in}
		\centering
		\includegraphics[width=1\linewidth]{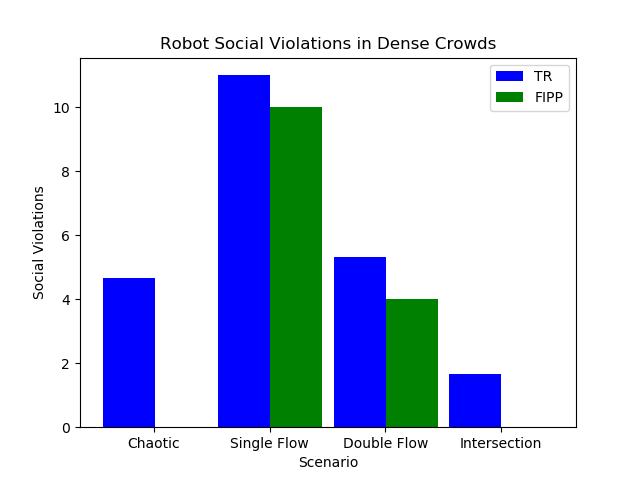}
		\subcaption{}
		\end{subfigure}
\caption{Results from simulation trials, a) the velocity of the robot across the trials and b) the social violations across the trials}
\label{fig:fsim}
\end{figure*}

\subsection{Flow-Informed Path Planning vs. Traditional Path Planning}

In order to evaluate the performance of FIPP we made use of SOCIALGYM, a lightweight 2D simulation environment for robot social navigation~\cite{art37}. The scenarios used for testing involved an empty space and between 25-50 randomly generated pedestrians with a specified flow. The robot was then randomly given a start and goal point. Four different flow Scenarios were created and tested on, a chaotic, a single flow, a double flow, and an intersection, figure ~\ref{fig:scenes} We compare FIPP to a trajectory-rollout (TR) local path planner in terms of efficiency (distance traveled / time)  and social acceptability. We evaluate social acceptability based on human proxemics~\cite{art37} in order to maintain human comfort the robot should remain in the social zone or at a distance of one to four meters. In the social zone two humans can maintain communicate without touching one another allowing for each person to have their own space. Distances between humans and robots less than one meter are known as the intimate zone and has the possibility of making humans uncomfortable~\cite{art42}. Thus, for ideal social compliance the robot would enter the intimate zone as little as possible. In our tests we define a social violation as the robot being firmly inside the intimate zone, thus anytime the robot comes within 0.5m of a human it is considered a social violation.

The simulation results show that in every scenario FIPP results in less social violations than does TR, sometimes navigating without any social violations at all. This ability to avoid social violation also often results in the robot being able to reach its goal faster than the traditional planner despite often having to take a longer route. This has to do with the number of social violations correlating to the number of times the robot has to stop to avoid a collision with a pedestrian, since this is largely unnecessary in the case of FIPP the robot is able to reach its goal faster than TR. Thus, FIPP is proven to be both more socially-compliant and more efficient than traditional path planners.

\subsection{Real-Robot Case Study}
\begin{figure}[h!]
\centering
\begin{subfigure}{1.5in}
    \centering
    \includegraphics[width=.95\textwidth]{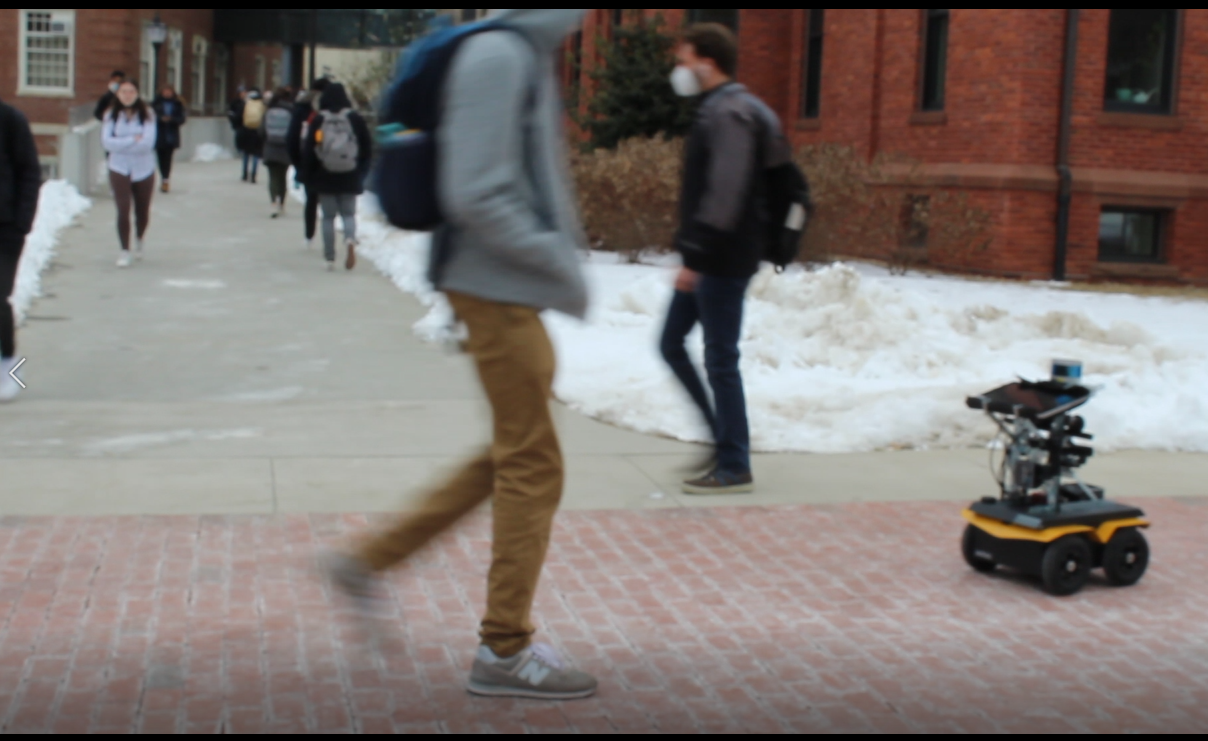}
    \subcaption{}
\end{subfigure}%
\begin{subfigure}{1.5in}
    \centering
    \includegraphics[width=1\textwidth]{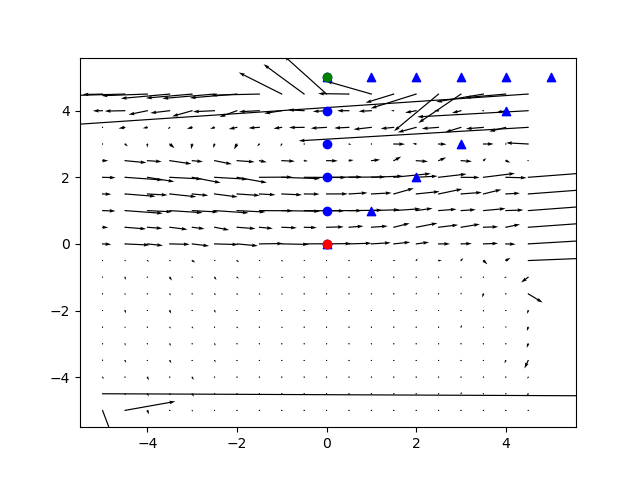}
    \subcaption{}
\end{subfigure}
\caption[short]{a) A photo of the scene where the real robot experiments were run. b) The flow field extracted from the scene, trajectory-rollout robot path (circles), FIPP robot path (squares), start and goal(red and green circles respectively) }
\label{fig:real}
\end{figure} 
In order to ensure that our method of Flow-Informed based path planning was valid on real-robots we employed a Clearpath Jackal and  ran a case study on the UMass Campus to compare our approach to that of trajectory-rollout with real pedestrians. A Clearpath Jackal equipped with a 3D LiDAR was used as the test platform. The trial was run during a passing period between classes where the flow of the pedestrians is both well established and reproducible several times a day. A* was used in place of LAO* in order to show the versatility of FIPP and its ability to find socially compliant solutions regardless of deterministic heuristic algorithm (figure ~\ref{fig:real}).

When run with trajectory-rollout the robot would take the shortest path to the goal disrupting pedestrians along its route. Pedestrians would often stop to allow the robot to pass in front of them and human intervention was necessary to keep the robot from colliding with pedestrians. All though efficient the robot displayed little social compliance when navigating this way. 

As opposed to the flow uninformed robot path planning, when running FIPP the robot was able to seamlessly enter two different flows in order to reach its goal. No human intervention was necessary and the social affect on the crowd was minimal. FIPP was able to allow the robot to navigate with social compliance and little to no affect of pedestrians.                           


\section{CONCLUSIONS}
In this paper we introduce FIPP, an efficient and socially compliant solution to navigation in densely crowded pedestrian environments. FIPP treats pedestrians as particles suspended in a fluid and extracts the flow of this fluid. FIPP then makes use of this flow to plan a path in and around dense crowds. We demonstrated FIPP's ability to navigate with greater social compliance and efficiency than traditional path planners both in simulation and on real robots. In future we plan to incorporate the inherent effect of the robots presence on the pedestrians flow to our flow-field models and to apply IRL to learning heuristics for FIPP.  








\begin{thebibliography}{10}

\bibitem{art41}
Lao\*: A heuristic search algorithm that finds solutions with loops.
\newblock {\em Artificial Intelligence}, 129(1):35--62, 2001.

\bibitem{art10}
Shreetam Behera, Debi~Prosad Dogra, Malay~Kumar Bandyopadhyay, and
  Partha~Pratim Roy.
\newblock Understanding crowd flow movements using active-langevin model.
\newblock {\em CoRR}, abs/2003.05626, 2020.

\bibitem{art48}
Pavan Bonkinpillewar, Ajinkya Kulkarni, Mahesh Panchagnula, and Srikanth
  Vedantam.
\newblock A novel coupled fluid–particle dem for simulating dense granular
  slurry dynamics.
\newblock {\em Granular Matter}, 17, 07 2015.

\bibitem{art28}
Yu~Fan Chen, Michael Everett, Miao Liu, and Jonathan~P. How.
\newblock Socially aware motion planning with deep reinforcement learning.
\newblock {\em CoRR}, abs/1703.08862, 2017.

\bibitem{art47}
Yu~Fan Chen, Miao Liu, Michael Everett, and Jonathan~P. How.
\newblock Decentralized non-communicating multiagent collision avoidance with
  deep reinforcement learning.
\newblock {\em CoRR}, abs/1609.07845, 2016.

\bibitem{art29}
Yuying Chen, Congcong Liu, Ming Liu, and Bertram~E. Shi.
\newblock Robot navigation in crowds by graph convolutional networks with
  attention learned from human gaze.
\newblock {\em CoRR}, abs/1909.10400, 2019.

\bibitem{art43}
Jiyu Cheng, Hu~Cheng, Max Q.-H. Meng, and Hong Zhang.
\newblock Autonomous navigation by mobile robots in human environments: A
  survey.
\newblock In {\em 2018 IEEE International Conference on Robotics and
  Biomimetics (ROBIO)}, pages 1981--1986, 2018.

\bibitem{art30}
Michael Everett, Yu~Fan Chen, and Jonathan~P. How.
\newblock Motion planning among dynamic, decision-making agents with deep
  reinforcement learning.
\newblock {\em CoRR}, abs/1805.01956, 2018.

\bibitem{art24}
Gonzalo Ferrer, Anaís Garrell, and Alberto Sanfeliu.
\newblock Robot companion: A social-force based approach with human
  awareness-navigation in crowded environments.
\newblock In {\em 2013 IEEE/RSJ International Conference on Intelligent Robots
  and Systems}, pages 1688--1694, 2013.

\bibitem{art37}
Edward~T. Hall.
\newblock {\em The hidden dimension.}
\newblock Doubleday, 1969.

\bibitem{art34}
Mahmoud Hamandi, Mike D'Arcy, and Pooyan Fazli.
\newblock Deepmotion: Learning to navigate like humans.
\newblock {\em CoRR}, abs/1803.03719, 2018.

\bibitem{art23}
Dirk Helbing and Péter Molnár.
\newblock Social force model for pedestrian dynamics.
\newblock {\em Physical Review E}, 51(5):4282–4286, May 1995.

\bibitem{art21}
L.F. Henderson.
\newblock On the fluid mechanics of human crowd motion.
\newblock {\em Transportation Research}, 8(6):509--515, 1974.

\bibitem{art20}
Roger~L. Hughes.
\newblock The flow of human crowds.
\newblock {\em Annual Review of Fluid Mechanics}, 35(1):169--182, 2003.

\bibitem{art42}
Helge H{\"u}ttenrauch, Kerstin~Severinson Eklundh, Anders Green, and Elin~Anna
  Topp.
\newblock Investigating spatial relationships in human-robot interactions.
\newblock 2005.

\bibitem{art31}
Beomjoon Kim and Joelle Pineau.
\newblock Socially adaptive path planning in human environments using inverse
  reinforcement learning.
\newblock {\em International Journal of Social Robotics}, 8:51--66, 2016.

\bibitem{art16}
Kenji Koide, Jun Miura, and Emanuele Menegatti.
\newblock A portable three-dimensional lidar-based system for long-term and
  wide-area people behavior measurement.
\newblock {\em International Journal of Advanced Robotic Systems}, 16, 02 2019.

\bibitem{art22}
Don~S. Lemons and Anthony Gythiel.
\newblock Paul langevin’s 1908 paper “on the theory of brownian motion”
  [“sur la théorie du mouvement brownien,” c. r. acad. sci. (paris) 146,
  530–533 (1908)].
\newblock {\em American Journal of Physics}, 65(11):1079--1081, 1997.

\bibitem{art45}
Eitan Marder-Eppstein, Eric Berger, Tully Foote, Brian Gerkey, and Kurt
  Konolige.
\newblock The office marathon: Robust navigation in an indoor office
  environment.
\newblock In {\em 2010 IEEE International Conference on Robotics and
  Automation}, pages 300--307, 2010.

\bibitem{art25}
Dhanvin Mehta, Gonzalo Ferrer, and Edwin Olson.
\newblock Autonomous navigation in dynamic social environments using
  multi-policy decision making.
\newblock In {\em 2016 IEEE/RSJ International Conference on Intelligent Robots
  and Systems (IROS)}, pages 1190--1197, 2016.

\bibitem{art39}
Jonathan Mumm and Bilge Mutlu.
\newblock Human-robot proxemics: Physical and psychological distancing in
  human-robot interaction.
\newblock {\em HRI: ACM SIGCHI/SIGART Human-Robot Interaction}, pages 331 --
  338, 2011.

\bibitem{art32}
N.~Perez-Higueras, F.~Caballero, and L.~Merino.
\newblock {Teaching Robot Navigation Behaviors to Optimal RRT Planners}.
\newblock {\em International Journal of Social Robotics}, 10(2):235--249, 2018.

\bibitem{art27}
No{\'{e}} P{\'{e}}rez{-}Higueras, Fernando Caballero, and Luis Merino.
\newblock Learning human-aware path planning with fully convolutional networks.
\newblock {\em CoRR}, abs/1803.00429, 2018.

\bibitem{art26}
Mike Phillips and Maxim Likhachev.
\newblock Sipp: Safe interval path planning for dynamic environments.
\newblock {\em 2011 IEEE International Conference on Robotics and Automation},
  pages 5628--5635, 2011.

\bibitem{art18}
Peter Trautman and Andreas Krause.
\newblock Unfreezing the robot: Navigation in dense, interacting crowds.
\newblock In {\em 2010 IEEE/RSJ International Conference on Intelligent Robots
  and Systems}, pages 797--803, 2010.

\bibitem{art44}
Rudolph Triebel, {Kai O.} Arras, Rachid Alami, Lucas Beyer, Stefan Breuers,
  Raja Chatila, Mohamed Chetouani, Daniel Cremers, Vanessa Evers, Michelangelo
  Fiore, {Omar A. Islas} Ram{\'i}rez, Michiel Joosse, Harmish Khambhaita,
  Tomasz Kucner, Bastian Leibe, {Achim J.} Lilienthal, Timm Linder, Manja
  Lohse, Martin Magnusson, Billy Okal, Luigi Palmieri, Umer Rafi, Marieke {van
  Rooij}, Hayley Hung, and Lu~Zhang.
\newblock {\em SPENCER: A socially aware service robot for passenger guidance
  and help in busy airports}, pages 607--622.
\newblock Springer Tracts in Advanced Robotics. March 2016.

\bibitem{art40}
Araceli Vega~Magro, Ramón Cintas, Luis Manso, Pablo Bustos, and Pedro Núñez.
\newblock {\em Socially-Accepted Path Planning for Robot Navigation Based on
  Social Interaction Spaces}, pages 644--655.
\newblock 01 2020.

\end{thebibliography}
\end{document}